# Seq2Seq AI Chatbot with Attention Mechanism

*(Final Year Project submitted in fulfillment of the requirements for the degree of master in Artificial Intelligence)*

**Abonia Sojasingarayar**        ABONIAA@GMAIL.COM

*Department of Artificial Intelligence*
*IA School/University-GEMA Group*
*Boulogne-Billancourt, France*

## ABSTRACT

Intelligent Conversational Agent development using Artificial Intelligence or Machine Learning technique is an interesting problem in the field of Natural Language Processing. In many research projects, they are using Artificial Intelligence, Machine Learning algorithms and Natural Language Processing techniques for developing conversation/dialogue agent. In the past, methods for constructing chatbot architectures have relied on hand-written rules and templates or simple statistical methods. With the rise of deep learning, these models were quickly replaced by end-to-end trainable neural networks around 2015. More specifically, the recurrent encoder-decoder model [Cho et al., 2014] dominates the task of conversational modeling. This architecture was adapted from the neural machine translation domain, where it performs extremely well. Since then a multitude of variations[Serbanetal.,2016]and features were presented that augment the quality of the conversation that chatbots are capable of[Richard.,2017].Among current chatbots, many are developed using rule-based techniques, simple machine learning algorithms or retrieval based techniques which do not generate good results. In this paper, I have developed a Seq2Seq AI Chatbot using modern-day techniques. For developing Seq2Seq AI Chatbot, We have implemented encoder-decoder attention mechanism architecture. This encoder-decoder is using Recurrent Neural Network with LSTM (Long-Short-Term-Memory) cells. These conversation agents are predominately used by businesses, government organizations and non-profit organizations. They are frequently deployed by financial organizations like bank, credit card companies, businesses like online retail stores and start-ups.

# 1 INTRODUCTION

Chatbot is a Software program that generates response based on given input to mimic human conversations in text or voice mode. These applications are designed to simulate human-human interactions. Their functioning can range from customer service, product suggestion, product inquiry to personal assistant

Recently, there has been a major increase in interest in the use and deployment of dialogue generation systems. Many major tech companies are using a virtual assistant or chat agent to ill the needs of Alex. Though they are primarily questioning answering systems, their adoption by major corporations has peaked interesting customers and seems promising for more advanced conversational agent system in research and development.

# 2. HISTORY AND RELATED WORK

## 2.1 EARLY APPROCH

Telegram released its bot API, providing an easy way for developers, to create bots by interacting with a bot, the Bot Father. Immediately people started creating abstractions in node.js, ruby and python, for building bots. Then I encountered another bot, Mitsuku which seemed quite intelligent. It is written in AIML (Artificial Intelligence Markup Language); an XML based "language" that lets developers write rules for the bot to follow. Basically, you write a PATTERN and a TEMPLATE, such that when the bot encounters that pattern in a sentence from user, it replies with one of the templates. We call this model of bots, Rule based model.
Rule based models make it easy for anyone to create a bot. But it is incredibly difficult to create a bot that answers complex queries. The pattern matching is kind of weak and hence, AIML based bots suffer when they encounter a sentence that doesn't contain any known patterns. Also, it is time consuming and takes a lot of effort to write the rules manually. What if we can build a bot that learns from existing conversations (between humans). This is where Machine Learning comes in. these models that automatically learn from data are called Intelligent models.

## 2.2 INTELLIGENT MODELS

The Intelligent models can be further classified into: Retrieval-based models and Generative models.

The Retrieval-based models pick a response from a collection of responses based on the query. It does not generate any new sentences, hence we don't need to worry about grammar.

The Generative models are quite intelligent. They generate a response, word by word based on the query. Due to this, the responses generated are prone to grammatical errors. These models are difficult to train, as they need to learn the proper sentence structure by themselves. However, once trained, the generative models outperform the retrieval-based models in terms of handling previously unseen queries and create an impression of talking with a human for the user.

## 2.3 THE ENCODER-DECODER MODEL

The main concept that differentiates rule-based and neural network based approaches is the presence of a learning algorithm in the latter case. An "encoder" RNN reads the source sentence and transforms it into a rich fixed-length vector representation, which in turn in used as the initial hidden state of a "decoder" RNN that generates the target sentence. Here, we propose to follow this elegant recipe, replacing the encoder RNN by a deep convolution neural network (CNN). … It is natural to use a CNN as an image "encoder", by first pre-training it for an image classification task and using the last hidden layer as an input to the RNN decoder that generates sentences [Oriol et al., 2014]

In this work, only deep learning methods applied to chatbots are discussed, since neural networks have been the backbone of conversational modeling and traditional machine learning methods are only rarely used as supplementary techniques.

## 2.4 RECURRENT NEURAL NETWORKS

A recurrent neural network (RNN) [Rumelhart et al., 1988] is a neural network that can take as input a variable length sequence $x = (x_1,...,x_n)$ and produce a sequence of hidden states $h = (h_1,...,h_n)$, by using recurrence. This is also called the unrolling or unfolding of the network, visualized in Figure 1. At each step the network takes as input $x_i$ and $h_{i-1}$ and generates a hidden state $h_i$. At each step $i$, the hidden state $h_i$ is updated by

$$h_i = f(Wh_{i-1} + Ux_i) \qquad (1)$$

where W and U are matrices containing the weights (parameters) of the network. f is a nonlinear activation function which can be the hyperbolic tangent function for example. The vanilla implementation of an RNN is rarely used, because it suffers from the vanishing gradient problem which makes it very hard to train [Hochreiter, 1998]. Usually long short-term memory (LSTM) [Hochreiter and Schmidhuber, 1997]or gated recurrent units(GRU) [Choet al., 2014]are used for the activation function. LSTMs were developed to combat the problem of long-term dependencies that vanilla RNNs face. As the number of steps of the unrolling increase it becomes increasingly hard for a simple RNN to learn to remember information seen multiple steps ago.

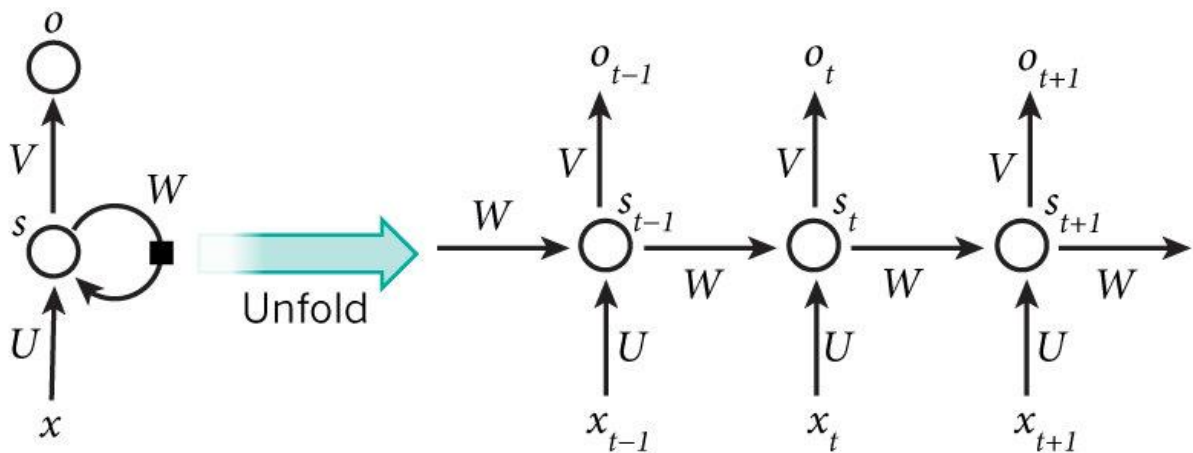

*Source:https://www.researchgate.net/figure/A-recurrent-neural-network-and-the-unfolding-in-time-of-the-computation-involved-in-its_fig1_324680970*

Figure 1: Unfolding of an RNN over 3 time-steps. Here x is the input sequence, o is the output sequence, s is the sequence of hidden states and U,W and V are the weights of the network. [Britz, 2015]

## 2.5 SEQ2SEQ MODEL

Sequence To Sequence model become the Go-To model for Dialogue Systems and Machine Translation. It consists of two RNNs (Recurrent Neural Network) , an Encoder and a Decoder. The encoder takes a sequence(sentence) as input and processes one symbol(word) at each time step. Its objective is to convert a sequence of symbols into a fixed size feature vector that encodes only the important information in the sequence while losing the unnecessary information. You can visualize data flow in the encoder along the time axis, as the flow of local information from one end of the sequence to another.
Each hidden state influences the next hidden state and the final hidden state can be seen as the summary of the sequence. This state is called the context or thought vector, as it represents the intention of the sequence. From the context, the decoder generates another sequence, one symbol (word) at a time. Here, at each time step, the decoder is influenced by the context and the previously generated symbols.

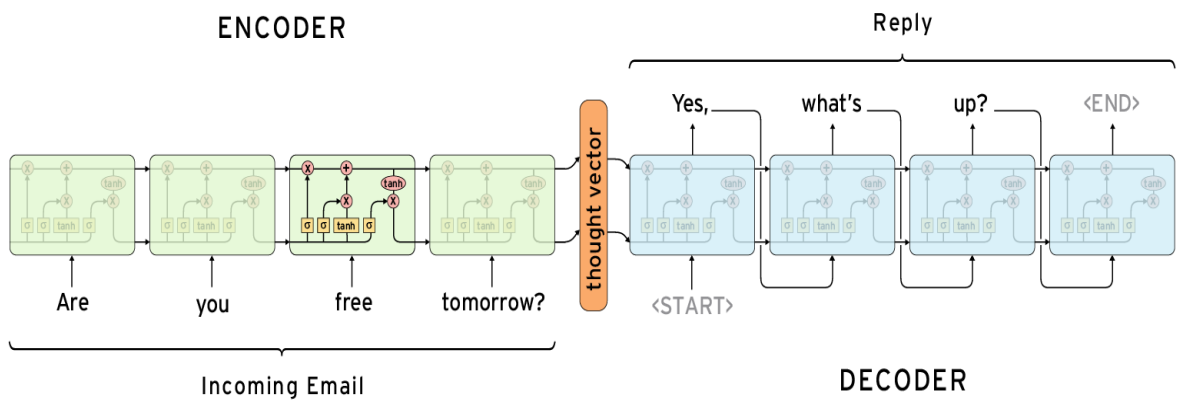

Figure 2: Long-Short-Term-Memory (LSTM) Encoder Decoder Architecture

There are a few challenges in using this model. The most disturbing one is that the model cannot handle variable length sequences. It is disturbing because almost all the sequence-to-sequence applications, involve variable length sequences. The next one is the vocabulary size. The decoder has to run softmax over a large vocabulary of say 20,000 words, for each word in the output. That is going to slow down the training process, even if your hardware is capable of handling it. Representation of words is of great importance. How do you represent the words in the sequence? Use of one-hot vectors means we need to deal with large sparse vectors due to large vocabulary and there is no semantic meaning to words encoded into one-hot vectors.

## 2.6 DECODING AND VOCABULARY

In order to get the actual generated output sequences there are several techniques to decode from the probabilities of the words. This is a greedy and deterministic approach, since it always outputs the same output for the same input. While decoding a word at each time-step is fine a better approach is to decode the whole output sequence at the same time, by outputting the sequence with the highest probability.

$$\hat{T} = \arg\max_{T} p(T|S)$$

*(2)*

Here S is the source sentence and T is the target sentence. Since in order to get the sequence with the highest probability, first all of the possible sequences have to be generated, a simple left-to-right beam search is usually employed to make the computation tractable. In the first time-step of the decoder, the top K words with the highest probabilities are kept. Then at each time-step this list is expanded by computing the joint probability of the partial sequences in the list and the words in the current time-step and retaining the K most probable partial sequences until the end of the output sequence is reached. [Richard., 2017]

Another important aspect of sequence-to-sequence models applied to tasks involving language is the vocabulary. The vocabulary consists of all the various words and symbols present in the dataset. One problem with this approach is that the vocabulary tends to be quite large, especially for very big datasets. Since the number of parameters of the model increases proportionally with the size of the vocabulary it is usually the case that the vocabulary is limited to some arbitrary size N. In this way only the embedding of the N most frequent words in the dataset are used and any other symbols are replaced with a common token representing unknown word. Many approaches have been proposed to the problem of handling out of vocabulary (OOV) or unknown words[Luong et al.,2014,Fengetal.,2017,Jean et al., 2014].

## 2.7    DEEP REINFORCEMENT LEARNING

"Deep Reinforcement Learning for Dialogue Generation"[2] Deep Reinforcement Learning (DRL) has been used for developing long conversation chatbots. Seq2Seq model can generate coherent dialogues but may produce repeated generic responses regardless of input and can get stuck in a loop in longer conversations. This occurs as Seq2Seq predicts utterances one at a time while ignoring their influence on future outcomes. Seq2Seq models tend to generate highly frequent repeated responses like "I don't know". This is due to the high frequency of generic responses in the training set, also these replies are more compatible with a wide range of input text.

| Baseline mutual information model (Li et al. 2015) | Proposed reinforcement learning model |
|---|---|
| A: Where are you going ? (1) | A: Where are you going ? (1) |
| B: I'm going to the restroom . (2) | B: I'm going to the police station . (2) |
| A: See you later . (3) | A: I'll come with you . (3) |
| B: See you later . (4) | B: no , no , no , no , you ?re not going anywhere . (4) |
| A: See you later . (5) | A: Why? (5) |
| B: See you later . (6) | B: I need you to stay here. (6) |
| ... | A: I don't know what you are talking about. (7) |
| ... | ... |
| A: how old are you ? (1) | A How old are you ? (1) |
| B: I'm 16 . (2) | B I'm 16. Why are you asking ? (2) |
| A: 16 ? (3) | A I thought you were 12. (3) |
| B: I don't know what you are talking about . (4) | B What made you think so ? (4) |
| A: you don't know what you are saying. (5) | A I don't know what you are talking about. (5) |
| B: I don't know what you are talking about . (6) | B You don't know what you are saying. (6) |
| A: you don't know what you are saying. (7) | ... |
| ... | ... |

*Source:https://www.researchgate.net/figure/Left-results-of-dialogue-simulation-between-two-conversational-agents-trained-on-the_fig10_313374979*

Figure 3: Left: results of dialogue simulation between two conversational agents trained on the Open subtitle dataset. Right: results of using additional reinforcement learning to keep the dialogue flow going for as long as possible. The first turn is always inputted manually. It is clear that the conversations on the right

side show more engagement and that the model with reinforcement learning achieves reasonably more "entertaining" dialogue.

In Dufarsky's paper, they have generated intermediate response using Seq2Seq model with attention where input was raw text. Then, the intermediate generated responses were fed into Reinforcement Model and was rewarded based on Ease of answering, Information Flow and Semantic Coherence. This is forward centric model, where if generated response is easy to answer, contribute to more information compared to previous dialogue history and grammatically and semantically correct, they are rewarded.

## 3 LIMITATIONS OF CHATBOTS

The blunt truth to know is, chatbots are machines, not people. Although you can attempt to equip them with a casual tone, chatbots will never truly sound human.
Chatbots are great at providing facts and data. But they can never truly create an emotional bond with customers. And building an emotional bond can be a make-or-break factor in today's competitive environment.
They often fail in long conversations and have reduced relevancy in dialogue generation. Most of this chatbots are developed for the restricted domain. The majority of them are using simple rule-based techniques. They perform well in question answering sessions and in very structured conversational modes. But, fail to emulate real human conversation and lacks flexibility in functioning.

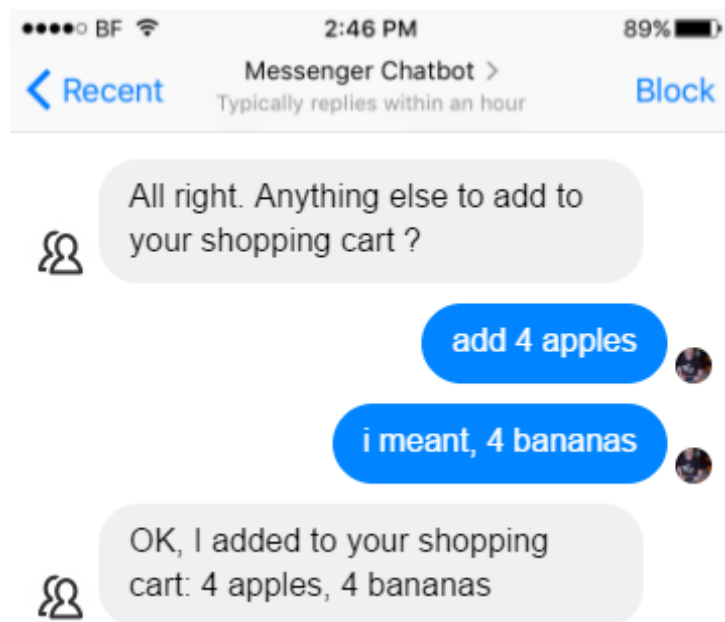

*Source: https://chatbotsmagazine.com/*
Figure 4: Encoder Decoder Architecture

A common misconception in the conversation design is to fail at recognizing user corrections. While intent and entity recognition work somehow well nowadays, thanks to evolving NLP platforms, most chatbot makers seem to "forget" to handle the intent "please correct what I said before."

## 4 ARCHITECTURE

In this project, I used tensorflow V1.14.0 to develop seq2seq model.Seq2Seq is industry standard choice for dialogue generation and many NLP tasks. It was coming from the tensorflow Seq2Seq contribution API. But also reasonably it is removed and no more accessible with recent tensorflow V2.X.But still the Seq2Seq API is accessible via tensorflow-addon package. Also to improve the model performance using state of art technique proposed by many research paper. In Seq2Seq models we can have options of attention mechanism, Beam search, bidirectionalRNN module.

### 4.1 LONG SHORT TERM MEMORY NETWORKS (LSTM)

Long Short Term Memory networks – usually just called "LSTMs" – are a special kind of RNN, capable of learning long-term dependencies. They were introduced by Hochreiter & Schmidhuber (1997).LSTMs are explicitly designed to avoid the long-term dependency problem. Remembering information for long periods of time is practically their default behavior, not something they struggle to learn.

All recurrent neural networks have the form of a chain of repeating modules of neural network. In standard RNNs, this repeating module will have a very simple structure, such as a single tanh layer. LSTMs also have this chain like structure, but the repeating module has a different structure. Instead of having a single neural network layer, there are four, interacting in a very special way as shown below.

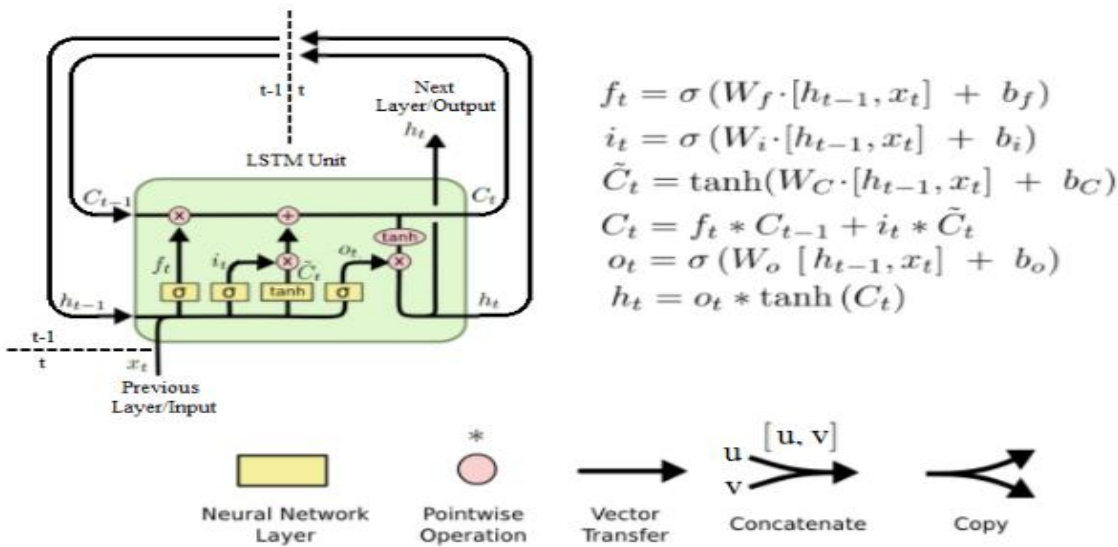

$x_t \in R^d$ : input vector to the LSTM unit
$f_t \in R^h$ : forget gate's activation vector
$i_t \in R^h$ : input gate's activation vector
$o_t \in R^h$ : output gate's activation vector
$h_t \in R^h$ : output vector of the LSTM unit
$c_t \in R^h$ : cell state vector

Figure 5: Long-Short-Term-Memory (LSTM)   and equation   *(3)*

## 4.2   ATTENTION MECHANISM

One of the limitations of seq2seq framework is that the entire information in the input sentence should be encoded into a fixed length vector, **context**. As the length of the sequence gets larger, we start losing considerable amount of information. This is why the basic seq2seq model doesn't work well in decoding large sequences. The attention mechanism, introduced in this paper, Neural Machine Translation by Jointly Learning to Align and Translate [2], allows the decoder to selectively look at the input sequence while decoding. This takes the pressure off the encoder to encode every useful information from the input.

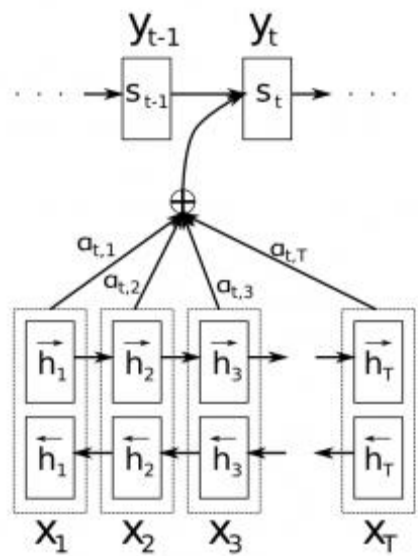

*Source: Neural Machine Translation by Jointly Learning to Align and Translate*

Figure 6: Attention Mechanism Architecture

During each time step in the decoder, instead of using a fixed context (last hidden state of encoder), a distinct context vector $c_i$ is used for generating word $y_i$. This context vector $c_i$ is basically the weighted sum of hidden states of the encoder.

$$c_i = \sum_{j=1}^{n} \alpha_{ij} h_j$$

(4)

where *n* is the length of input sequence, hj is the hidden state at time step *j*.

$$\alpha_{ij} = \exp(e_{ij}) / \sum_{k=1}^{n} \exp(e_{ik})$$

(5)

eij is the alignment model which is function of decoder's previous hidden state si−1 and the jth hidden state of the encoder. This alignment model is parameterized as a feed forward neural network which is jointly trained with the rest of model.

Each hidden state in the encoder encodes information about the local context in that part of the sentence. As data flows from word 0 to word n, this local context information gets diluted. This makes it necessary for the decoder to peak through the encoder, to know the local contexts. Different parts of input sequence contain information necessary for generating different parts of the output sequence. In other words, each word in the output sequence is aligned to different parts of the input sequence. The alignment model gives us a measure of how well the output at position *i* match with inputs at around position *j*. Based on which, we take a weighted sum of the input contexts (hidden states) to generate each word in the output sequence.

To build state-of-the-art neural machine translation systems, the attention mechanism works well, which was first introduced in 2015. The idea of the attention mechanism is to form direct short-cut connections between the target and the source by paying "attention" to relevant source content as we translate.[3]

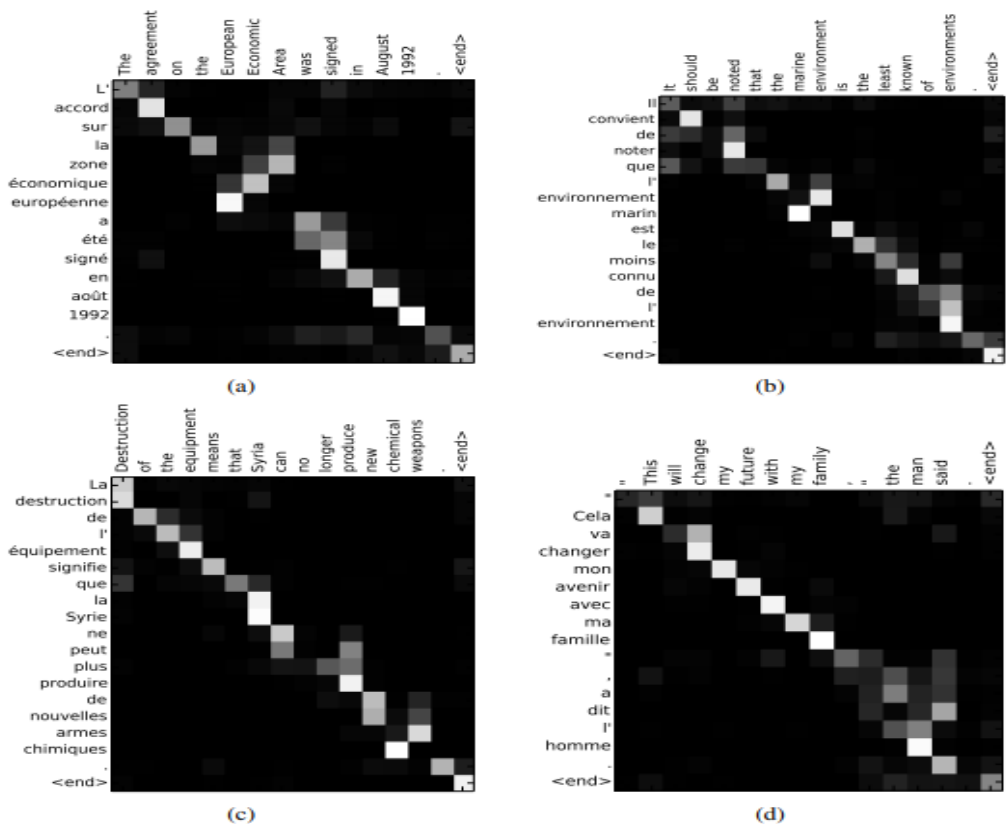

*Source: Neural Machine Translation by Jointly Learning to Align and Translate*

Figure 7: Attention visualization–example of the alignments between source and target sentences.

## 5 DATASET

### 5.1 DATA COLLECTION

We are going to use popular movie subtitle corpus which is"Cornell movie subtitle corpus". This corpus contains a metadata-rich large collection of conversations extracted from raw movie scripts from popular movies.

The following are found in the corpus:
- 220,579 conversational exchanges between 10,292 pairs of movie characters.
- Involves 9,035 characters from 617 movies
- In total 304,713 utterances

Other movie meta-data included genres, release year, IMDB rating, number of IMDB votes, IMDB rating.

## 5.2 DATA PREPROCESSING

Conversation data in the movie corpus contained Movie ID, Character ID, and Movie Line ID was separated by"+++++".

For preprocessing, conversation data were cleaned to remove this meta-data (eg. movie ID, character ID, Line ID). Also, data separators ("+++++") were eliminated. Additionally, some of the characters in the data contained an unsupported encoding format by UTF-8 standard and hence was removed.

Finally, data were separated into two different lists to assimilate with the format of Sequence to Sequence model (Seq2Seq) model input pipeline format where first list is the dialogue 1(or questions) and the second one was the response to dialogue 1(or answer).

After separating the two lists, data in both was cleaned simultaneously. Everything except alphabetical character and some punctuation (. , ?!') was removed as they hold little meaning in conversation. Also, all the text was converted to lowercase. Then, multiple consequent occurrences of this punctuation (. , ?!') was reduced to one in order to reduce punctuation overload. Next, all the punctuation except (') was separated with a single space before and after for better performance in the Sequence to Sequence model (Seq2Seq) module. Finally, all the consequent multiple spaces were reduced to single space and each text string was trimmed to remove before and after space. Also, data was cleaned for removing extraneous dialogues. If multiple consequent utterances from a single person were present everything except the last utterance for the person was stored. Filter the question and answers that are too short or long (here I used 2 as my minimum length and 5 as my maximum length) i.e. the sentence with 2 to 5 words are taken into account. The most frequent 6282 words in the training data were kept as vocabulary. Additionally the <PAD> token was used for padding input sequences to same lengths, the <EOS> token was used to signal the end of an utterance and the <UNK> token was used to replace all words not present in the vocabulary. <GO> was given to the start of the sentence. In total this gives a vocabulary size of 6286.

Before training, we work on the dataset to convert the variable length sequences into fixed length sequences, by padding. We use a few special symbols to fill in the sequence. Consider the following query-response pair.
Q: How are you?
A: I am fine.
Assuming that we would like our sentences (queries and responses) to be of fixed length, 10, this pair will be converted to:

    Q: [PAD, PAD, PAD, PAD, PAD, PAD, "?", "you", "are", "How"]
    A: [GO, "I", "am", "fine", ".", EOS, PAD, PAD, PAD, PAD]

Introduction of padding did solve the problem of variable length sequences, but consider the case of large sentences. If the largest sentence in our dataset is of length 100, we need to encode all our sentences to be of length 100, in order to not lose any words. Now, what happens to "How are you?"? There will be 97 PAD symbols in the encoded version of the sentence. This will overshadow the actual information in the sentence.

Bucketing kind of solves this problem, by putting sentences into buckets of different sizes. Consider this list of buckets: [(5, 10), (10, 15), (20, 25), (40, 50)]. If the length of a query is 4 and the length of its response is 4 (as in our previous example), we put this sentence in the bucket (5, 10). The query will be padded to length 5 and the response will be padded to length 10. While running the model (training or predicting), we use a different model for each bucket, compatible with the lengths of query and response. All these models share the same parameters and hence function exactly the same way. If we are using the bucket (5, 10), our sentences will be encoded to:

    Q: [PAD, "?", "you", "are", "How"]
    A: [GO, "I", "am", "fine", ".", EOS, PAD, PAD, PAD, PAD]

Word Embedding is a technique for learning dense representation of words in a low dimensional vector space. Each word can be seen as a point in this space, represented by a fixed length vector. Semantic relations between words are captured by this technique. The word vectors have some interesting properties.

    Paris – France + Poland = Warsaw.

The vector difference between Paris and France captures the concept of capital city. Word Embedding is typically done in the first layer of the network: Embedding layer, that maps a word (index to word in vocabulary) from vocabulary to a dense vector of given size. In the seq2seq model, the weights of the embedding layer are jointly trained with the other parameters of the model.

## 6  IMPLIMENTATION DETAILS

Algorithm:

| Algorithm | Deep Neural Network (DNN), Recurrent Neural Network (RNN) |
|---|---|
| Main Technique | Sequence to Sequence (Seq2seq) modeling with encoder |
| decoder Enhancement Techniques | Long Short Term Memory (LSTM) based RNN cell, Bidirectional LSTM, Neural attention Mechanism |

Web Application:

| Frond and backend technology used | HTML,CSS,Javascript Python Flask app |
|---|---|

Table 1: (a) Algorithm and technique used in this experiment.(b)Front and backend technologies used to build the web application in order to provide user interface

## 7 CONFIGURATION

As shown below I have tried three configurations for the model and config 3 outperform the other two configurations as it is more deep seq2seq model.

| Parameters | Config 1 | Config 2 | Config3 |
|---|---|---|---|
| batch size | 128 | 512 | 32 |
| embedding size | 128 | 512 | 1024 |
| Rnn size | 128 | 512 | 1024 |
| learning_rate | 0.001 | 0.001 | 0.001 |
| epochs | 500 | 100 | 50 |
| keep probability | 0.75 | 0.75 | 0.7 |
| min_learning_rate | 0.0001 | 0.0001 | 0.0001 |
| learning_rate_decay | 0.9 | 0.9 | 0.9 |

Table 2: Configuration table shows the different combination of hyperparameters used to improve model performance during experiment.

## 8 HARDWARE SPECIFICATION

Training in my local machine on CPU with the follow specification:

| Processor : | 8th Gen Intel Core i5-8250U CPU@1.60GHz |
|---|---|
| Ram : | 8GB. |

Table 3: Hardware Specification (Local machine)

Training in Google Colab platform using GPU took relatively less time.So I used Google Colab for config2.

## 9 TRAINING MODEL:

As we created the filtered question and answer list, we create training data (length of 22992) and validation data (I took the length of batch size).

As I trained the model for three different conversation as we discussed before using Adam optimizer as it can computes individual adaptive learning rates for different parameters from estimates of first and second moments of the gradient. We used BidirectionalLSTM is used in encoder side and attention mechanism is used in decoder side to improve model performance.

## 10 RESULT

In this section a detailed analysis of the trained models is given. They are also compared with a baseline seq2seq model.

| Model with Config1 | Model with Config2 |
| --- | --- |
| Human :hey | Human :hey |
| Bot :hi | Bot :hi |
| Human :How are you | Human :what doing |
| Bot :fine | Bot :sleep |
| Human :when will you be okay | Human :when will you be okay |
| Bot :okay | Bot :just few days |
| Human :where are you from | Human :where are you from |
| Bot : Sorry, I dint understand your context | Bot :southern California |
| Human :who am i | Human: you want to come with me? |
| Bot :yes | Bot: how is that for you? |
| Human: ok? | Human :how much do you like me |
| Bot : Sorry, I dint understand your context | Bot :you are not cold mostly |
| Human :what about weather in NY | Human :who am i |
| Bot : Sorry, I dint understand your context | Bot :doctor Livingston |
| Human :What's your hobby | Human :what's your hobby |
| Bot : going | Bot :eat |
| Human :omg | Human :wow you are best |
| Bot : Sorry, I dint understand your context | Bot :I have done |
| Human :what you know | Human :really well |
| Bot :I know | Bot :goodbye |

Table 4: Some interesting responses from bot for different configuration/model.Left side shows the responses of model trained with config1[Table 2] and right side shows the responses of model trained with config2[Table 2].Human indicates the user text and Bot indicate the reply we get from the model.

## 11 USER INTERFACE

GUI (graphical user interface) was developed using flask app. Flask app that provides a chat interface to the user, to interact with our seq2seq model. The reply method in the code is called via an AJAX request from index.js.It sends the text from user to our seq2seq model, which returns a reply. The reply is passed to index.js and rendered as text in the chat box.

In below figure orange line text is reply from bot and blue line text is reply from human

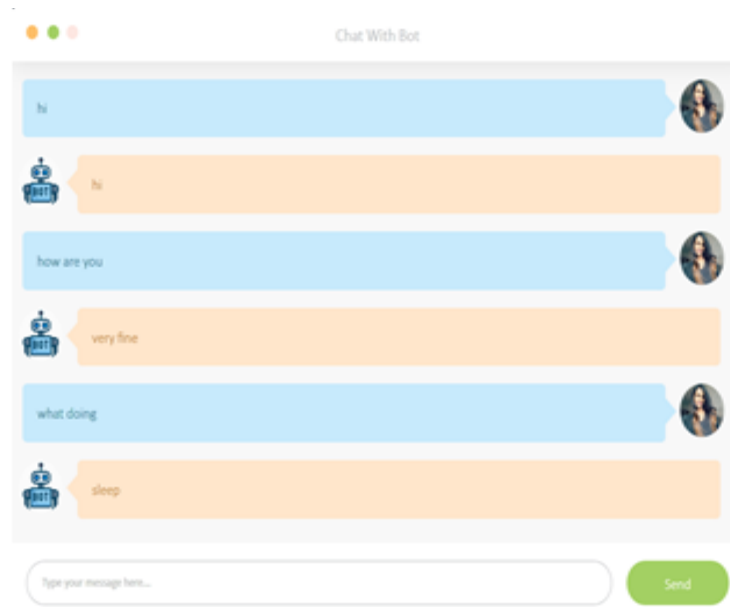

Figure 8: Web Application/Chat Interface

## 12  CHALLENGES

Training is a long process which demands higher processing power and configured computing machine. Another problem is finding right hyper parameters to optimize our model.

Developing a generic chatbot is really challenging. The model used in this experiment is for machine translation, the dialogue generation is treated as translation problem, where histories of earlier conversations are not taken into account. Hence, the model can be limited in performance regarding long conversation. Many of the output were repetitive and generic. Also, due to lack of real-life quality data the chatbot performed somehow below optimum for imitating human interaction. Also, many utterances were discarded due to longer length or discrepancy.

## 13  FUTURE WORK

Use attention mechanism like luong attention which is also suggested in many papers. Also can try different hyperparameters and evaluation metrics to improve the chat bot performance.

## 14   CONCLUSION

Various techniques and architectures were discussed, that were proposed to augment the encoder-decoder model and to make conversational agents more natural and human-like. Criticism was also presented regarding some of the properties of current chatbot models and it has been shown how and why several of the techniques currently employed are inappropriate for the task of modeling conversations. The performance of the training was analyzed with the help of automatic evaluation metrics and by comparing output responses for a set of source utterances. As the cornel data must be improved further to get better results in future. We can also try different combination of hyperparameters other than the configurations discussed in this paper. The training on Cornell Movie Subtitle corpus produced result which needs further improvement and more attention on training parameters. We can try different attention mechanism like Luong.We can also judge the quality of dataset by using another similar dataset with the hyperparameters we tried for cornel data.